\documentclass[journal, twoside, web]{IEEEtran}

\usepackage{cite}
\usepackage{amsmath,amssymb,amsfonts}
\usepackage{algorithmic}
\usepackage{algorithm}
\usepackage{graphicx}
\usepackage{textcomp}
\usepackage{tikz}
\usepackage{xcolor}
\usepackage{hyperref}

\usepackage{amsmath,amssymb} 
\usepackage{physics} 
\usepackage{xspace} 
\usepackage{bm} 
\usetikzlibrary{positioning} 
\usetikzlibrary{arrows.meta} 
\usetikzlibrary{calc} 
\tikzset{>=latex} 

\begin{document}
\title{NCDD: Nearest Centroid Distance Deficit for Out-Of-Distribution Detection in Gastrointestinal Vision}
\author{Sandesh Pokhrel*, Sanjay Bhandari*, Sharib Ali, Tryphon Lambrou, Anh Nguyen, Yash Raj Shrestha, Angus Watson, Danail Stoyanov, Prashnna Gyawali, Binod Bhattarai
\thanks{* denotes authors with equal contribution}
\thanks{Sandesh P. and Sanjay B. are affiliated with the Nepal Applied Informatics and Mathematics Institute for Research, Nepal. Sharib Ali is a lecturer at the University of Leeds, UK. Similarly, Anh Nguyen is a Senior Lecturer (Associate Professor) at the University of Liverpool, UK. Yash R. Shrestha is an Assistant Professor at the University of Lausanne, Switzerland. Danail Stoyanov is a Professor at University College London (UCL), London. Prashnna Gyawali is an Assistant Professor at West Virginia University, USA.
Tryphon Lambrou is an Associate Professor, Angus Watson is a Clinical Professor of Surgery, and Binod Bhattarai is an Assistant Professor, all at the University of Aberdeen, UK.
}
}

\maketitle

\begin{abstract}
The integration of deep learning tools in gastrointestinal vision holds the potential for significant advancements in diagnosis, treatment, and overall patient care.  A major challenge, however, is these tools' tendency to make overconfident predictions, even when encountering unseen or newly emerging disease patterns, undermining their reliability. 

We address this critical issue of reliability by
framing it as an out-of-distribution (OOD) detection problem, where previously unseen and emerging diseases are identified as OOD examples.
However, gastrointestinal images pose a unique challenge
due to the overlapping feature representations between in-
Distribution (ID) and OOD examples. Existing approaches often overlook this characteristic, as they are primarily developed for natural image datasets,
where feature distinctions are more apparent. Despite the overlap, we hypothesize that the features of an in-distribution example will cluster closer to the centroids of their
ground truth class, resulting in a shorter distance to the
nearest centroid. In contrast, OOD examples maintain 
an equal distance from all class centroids. Based on this observation, we propose a novel nearest-centroid
distance deficit (NCCD) score in the feature space for gastrointestinal OOD detection.

Evaluations across multiple deep learning architectures and two publicly available benchmarks, Kvasir2 and Gastrovision, 
demonstrate the effectiveness of our approach compared to several state-of-the-art methods. The code and implementation details are publicly available at: \href{https://github.com/bhattarailab/NCDD}{\textcolor{blue}{https://github.com/bhattarailab/NCDD}}
\end{abstract}


\section{Introduction}
\label{sec:introduction}
\IEEEPARstart{G}{astrointestinal} diseases are among the most prevalent worldwide, with over seven billion incidents and 2.8 million cases reported in 2019 \cite{DigestiveDiseaseBurden}. 38.4\% of the prevalent diseases in the same year had digestive etiology, with a significant portion being gastrointestinal disorders \cite{percentageofcases}. A more concerning fact is that even with the advancement of technology, the number of deaths and crude deaths per population increased significantly from 2000 to 2019 on account of these diseases. 

A relatively common method of diagnosing gastrointestinal diseases is endoscopy, which is used to detect and diagnose any abnormalities present in the gastrointestinal tract with the help of a light pipe. With growing cases, the need for fast and accurate diagnosis has increased, leading to technological advancements like capsule endoscopy. However, manually diagnosing such cases has become increasingly challenging \cite{endoscopyguide} for practitioners and requires aid through automated diagnosis. Deep learning has shown significant promise in this regard. Especially, CNNs have been used to recognize anatomical landmarks seen in the endoscope and have also been effective in detecting and segmenting abnormalities including but not limited to gastric cancer \cite{gastriccancerAICNN}, polyps, ulcerative colitis, and esophagitis \cite{svmcnngastrodisease,gitractensemblestacking,dlendodisease}. 

Although deep learning algorithms perform fairly well in close-set classification tasks, such as distinguishing anatomical landmarks and well-studied abnormalities, they fail to do so in open-set classification tasks, such as detecting rare and novel anomalies. These models do not account for rare diseases and unseen examples of abnormalities properly, with sparse data examples to train on leading to over-confident prediction of unknown cases questioning their overall trustworthiness and reliability \cite{oodmedicalapplications}. 

In existing methods, abnormality detection in the GI tract has been modeled as a closed-set classification problem and can recognize only a known abnormality \cite{giclassification, gidlclassification}. In some scenarios, the search for abnormalities in gastrointestinal vision is framed as a detection problem. Different cases of abnormalities are categorized as objects on which the model is trained to learn features and make predictions on unseen images \cite{AnamolyasObjectdet}. 
These classification and detection-based models are trained on a fixed set of seen abnormalities, and thus, the expectation of discovering and flagging rare, unseen, or emerging diseases, which pertain more to novelty detection, is not feasible with these models. Furthermore, the resources required to annotate different types of abnormalities, including rare cases, are challenging. 

\begin{figure}
\includegraphics[width=\linewidth]{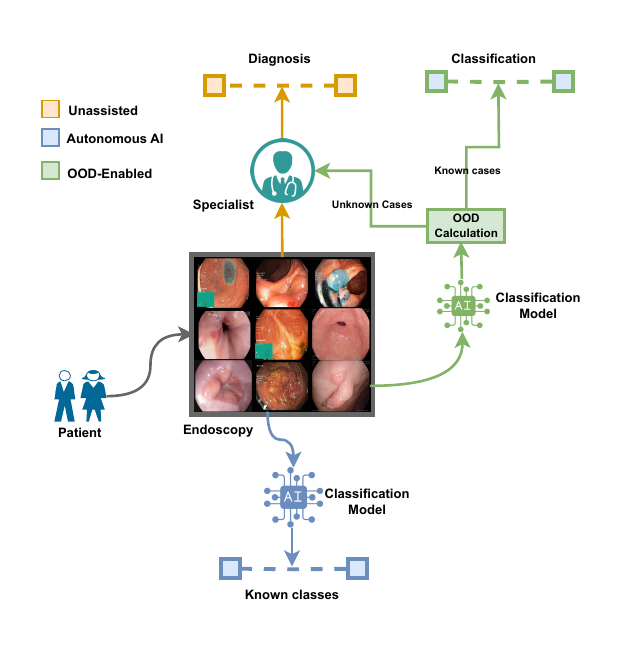}
\centering
\caption{Landscape of clinical procedures in gastrointestinal vision. Orange: Unassisted, a doctor has to assess all patients' data tediously and redundantly. Blue: Artificial Intelligence can help in the classification of known or seen diseases but makes misleading assumptions and often overconfident predictions on images when it faces real-world examples consisting of examples that it had never seen. Green: A combination of human intervention and OOD enabled the AI method to improve efficacy in the current scenario, where a specialist intervenes to correct any unseen or unknown instances that the AI model is uncertain in classifying.} \label{clinical_view}
\end{figure}

For robust identification of abnormalities in gastrointestinal images and to make this process reliable, we formulate the process as an out-of-distribution detection (OOD) problem. OOD is a popular problem in computer vision ~\cite{oodsurvey,oodmedicalapplications} and natural language processing~\cite{OODNLP}. However, it has received little attention in the medical domain, limited to magnetic resonance imaging and computed tomography~\cite{MOOD,optical_tomography,coopd}, ultrasounds~\cite{dualconditioneddiffusion}. OOD enables us to identify the close-set categories and flag any example different from the close-set examples, also called in-domain (ID) examples. This characterization is extremely useful for making AI algorithms reliable, rather than making unnecessary wrong predictions on unseen and rare examples, and providing opportunities for human intervention. Moreover, such features are essential in the medical domain where a single mistake in the algorithm can cost someone's life. By integrating clinicians in the loop for the cases where an algorithm is uncertain in its decisions, we can move towards trustworthiness in the practical application of these algorithms.

\begin{figure}[b!]
\includegraphics[width=\linewidth]{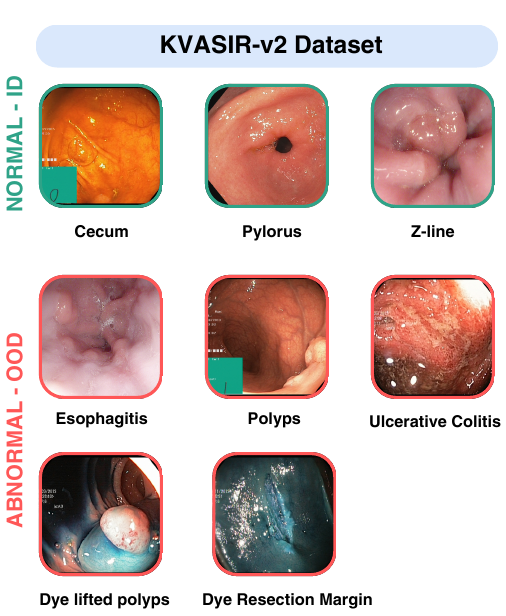}
\centering
\caption{The Kvasirv2 dataset \cite{kvasir} is formulated for OOD detection of abnormalities. Three classes, Z-line, Cecum, and Pylorus, are healthy cases showing normal anatomical landmarks in the dataset, while the remaining are abnormalities, either pathological conditions or images seen during the treatment procedure.}
\label{fig1}
\end{figure}

In this paper, we tackle the challenge of developing reliable learning algorithms for gastrointestinal vision by introducing a novel OOD detection framework. In our study, ID examples are healthy anatomical landmarks of the GI tract such as Z-line, Cecum, and Pylorus, as shown in Figure \ref{fig1} (top). In contrast, OOD examples correspond to any abnormalities in these landmarks, as illustrated in Figure \ref{fig1} (bottom). We take a supervised learning approach to classify normal anatomical landmarks in gastrointestinal endoscopic images while simultaneously detecting any abnormalities as OOD cases. 

Towards this, we combine the information about the proximity of an image in feature space to the nearest ID class centroid and information about how far it
is from non-nearest ID class centroids. Without loss of generality, the minimization of loss function for classification of 
ID examples minimize intra-class distance while maximizing inter-class distance, making the representations of images belonging to a class more compact. In contrast, the model is suboptimal for OOD examples, causing them to be more scattered in the feature space. 
This forms the basis of our proposed OOD detection method, the Nearest Centroid Distance Deficit (NCDD) score, which outperforms existing methods for the OOD detection tasks in the gastrointestinal domain.

Our specific contributions can be summarized as follows:
\begin{itemize}
\renewcommand\labelitemi{{\normalfont \bfseries $\bullet$}} 
    \item We formulate abnormality detection as an OOD problem for models trained on the classification of normal anatomical landmarks, allowing us to detect the presence of pathologies in endoscopic images without the need for the model to be explicitly trained on it.
    \item We introduce a novel distance-based OOD detection method, NCDD, that utilizes the information from the nearest-class cluster and the non-nearest-class cluster to distinguish ID from OOD. 
    \item Our method is easy-to-use, and post-hoc method. This means it's straightforward to implement and can be applied across various model architectures.
\end{itemize}

Our method is evaluated on two different gastrointestinal datasets and four separate classifier backbones, ResNet-18\cite{resnet}, ViT\cite{vit}, DEiT\cite{deit} and MLP-Mixer\cite{deit}, demonstrating its superior performance compared to existing methods.

\section{Literature Review}

Deep latent spaces and embedding clustering-based methods have been used for some time for out-of-distribution (OOD) detection. Dinari \textit{et al.}~\cite{dinari2022variational} trained a conditional variational model with Kullback-Leibler loss, a triplet loss, and a distancing loss that pushes classes away from each other to improve class separation in the latent space. During inference, the class-dependent log-likelihood values of a deep feature ensemble of the test point are also weighted based on reconstruction errors, further improving the decision rule. Another approach \cite{sundar2020out} focused on multi-label datasets, utilizing \(\beta\)-VAEs trained on partitioned data to identify sensitive latent variables for detecting OOD samples concerning specific generative factors. Sinhamahapatra \textit{et al.} \cite{sinhamahapatra2022all} investigated clustering in the embedding space by applying k-means/Gaussian Mixture Model (GMM) clustering on the learned embeddings, utilizing distance metrics and probability scores derived from cluster distributions to perform OOD detection.
Apart from deep latent spaces and clustering-based methods, many logits-based \cite{energyscore,msp,maxlogit}, probability-based \cite{entropy}, gradient-based \cite{Odin}, and feature space-based \cite{knn-ood} approaches have been popular in the field of OOD detection literature and have also proved promising.
Hendricks \textit{et al.} \cite{msp} proposed MSP, using maximum predicted softmax probability as the ID score. The Entropy method \cite{entropy} sums over the entropy of predicted probabilities. The Maximum Logit method \cite{maxlogit}, also by Hendricks \textit{et al.}, is effective even in multi-class settings. The Energy score \cite{energyscore} uses the $logsumexp$ of the logits to distinguish ID from OOD. Odin \cite{Odin} uses probability and gradient information to generate an OOD score. 

The non-parametric KNN-OOD method \cite{knn-ood} uses the distance between a sample image in feature space and the $k$-th nearest neighbour from the training set to determine ID and OOD samples. More recently, methods such as BLOOD\cite{BLOOD} have looked into the gradient information within separate layers of the network to separate OOD from ID data. Another recent OOD method, Neco~\cite{neco}, exploits the geometric properties of neural collapse and the principal component spaces for OOD detection. In making OOD detection fast as well as effective, Liu. \textit{et al.} proposed FDBD\cite{fdbdood}, wherein they utilize the rich feature information to generate a decision boundary with the help of which they can mark data points sufficiently away from the boundary as OOD. Although these methods are used in generalized real-world settings of computer vision, are not studied in medical scenarios, and fewer have been tested in gastrovision. 

In medical deep learning, out-of-distribution (OOD) detection has attracted much attention recently in detecting abnormality as OOD \cite{dualconditioneddiffusion,coopd} rather than as a class. Arnau \textit{et al.} \cite{selfsupendo} showed that self-supervised learning in capsule endoscopy could help to discern abnormalities and healthy sections of the abdomen, but there are no explorations on the supervised front for gastrointestinal OOD. COOpD \cite{coopd} formulated abnormalities as OOD to separate homogeneous healthy regions from heterogeneous pathological images of chest X-rays. Mehta \textit{et al.} ~\cite{oodskinlesion} proposed a mixup strategy to correct the long-tail data problem in realistic settings of a skin lesion for OOD detection. The need for OOD algorithms has been made evident by various cases of strong and well-trained models failing in out-of-the-wild examples, which often have to be dealt with in real-world scenarios \cite{oodsurvey}. The importance of being able to flag OODs in medical scenarios is especially crucial as they support experts in detecting incidental findings that could otherwise be overlooked\cite{MOOD}. 

Despite this, there is a lack of study on how well existing OOD scoring methods perform on gastrointestinal images, especially when trained only on anatomical landmarks and normal findings, and whether these models can be used to identify abnormalities. To this end, our method reinforces trustworthiness in gastrointestinal vision through a novel NCDD method, which utilises non-nearest class centroid information from the conventional classification method to effectively discern abnormalities as OOD.

\section{Method}

\begin{figure*}[!ht]
\includegraphics[width=\linewidth]{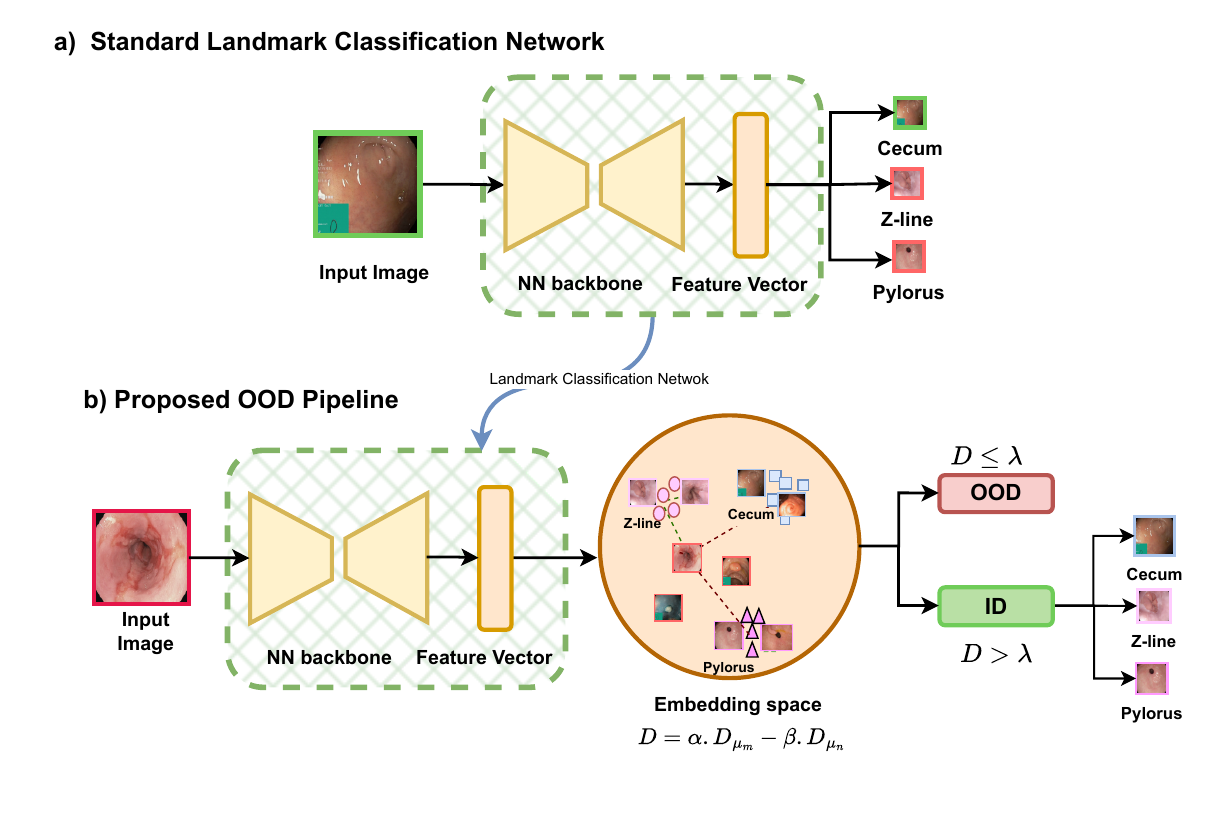}
\centering
\label{pipelinefigure}
\caption{\textbf{a)} Standard classification pipeline in landmark classification for Kvasirv2 \cite{kvasir} in-distribution dataset. In this pipeline, images fed as input will always be one of the three classes, regardless of how unrelated they are to the model's capability. In essence, the model doesn't know whether it is capable of making inferences on an image or not. \textbf{b)} Overview of our proposed OOD detection method: based on the feature representation distances for any given image, we can know whether the image is something the model has knowledge of or not.} \label{pipelinefig}
\end{figure*}

\subsection{Preliminaries: Out-of-Distribution Detection}
\subsubsection{Setup}
We consider supervised multi-class classification, where $X$ is the input space, $Y$ = $\{1, 2, \ldots, C\}$ is the label space and $P_{XY}$ is the distribution over $X\times Y$. Let $P_{in}$ be the marginal distribution of $X$ which is comprised of the healthy person data (In-Distribution).

\subsubsection{Supervised Learning}
For supervised learning, a neural network $f: {X} \rightarrow \mathbb{R}^{|Y|}$, is trained on i.i.d samples extracted from $P_{X, Y}$, to minimize the loss function \textit{L} over the input dataset. Here, $f: {X} \rightarrow \mathbb{R}^{|Y|}$ is the classification predicted for input $X$ by the model $f$. The model $f$ consists of encoder $\varphi$ and a fully connected layer $FC$. Feature vector $z$ : $\varphi(x)$  is produced from the encoder $\varphi$.

\subsubsection{OOD Detection}
At test time, the trained network can be presented input from a distribution $P_{out}$, quite different from $P_{in}$. The main objective of the out-of-distribution detection is to differentiate whether the sample belongs to $P_{in}$ or $P_{out}$. Our work reformulates out-of-distribution detection as binary classification problem where a decision function $g(x)$ is defined by the scores produced by the scoring function $SC(x)$ and a threshold $\lambda$, as given by:

\begin{equation} \label{eq-ood}
\begin{split}
g(x) = & \left\{
\begin{array}{ll}
\textit{OOD}, & \text{if } \text{score} \leq \lambda \\
\textit{ID}, & \text{otherwise}
\end{array}
\right. \\
\end{split}
\end{equation}

The threshold $\lambda$ is usually chosen to have a
true positive rate of 95\% over the input dataset.  Figure \ref{pipelinefig} compares our method with the vanilla classification method. 
%
%

Given that our problem is based on a classification problem, as illustrated in Figure~\ref{pipelinefig}, we employ the cross-entropy loss function to learn the model parameters. The cross-entropy loss function is defined as:
\begin{equation}
L = -\sum_{i=1}^C y_i \log(\hat{y}_i)
\end{equation}
where:
\begin{itemize}
\item $C$ is the number of classes
\item $y_i$ is the true probability of class $i$ (usually 0 or 1 for hard labels)
\item $\hat{y}_i$ is the predicted probability of class $i$
\end{itemize}

While our approach uses cross-entropy loss, it can be adapted to other objective functions suitable for close-set classification problems. Cross-entropy loss influences the feature space by enabling the model to assign high probabilities ($\hat{y}_i$ close to 1) to the correct class ($y_i$) and low probabilities to incorrect classes during training. As the loss minimizes, the feature space evolves, pushing examples from different  ID classes further apart, creating distinct clusters, and separating their corresponding centroids. Simultaneously, the data points within the same classes are compressed, resulting in tighter clusters around their respective centroids~\cite{separabilityCE}, which is depicted in the embedding space in Figure~\ref{pipelinefigure}. Based on this clustering behaviour induced by cross-entropy loss, we propose an out-of-distribution (OOD) detection method that examines how OOD data positions itself relative to these class centroids in feature space. 

\begin{figure}[!ht]
\includegraphics[width=\linewidth]{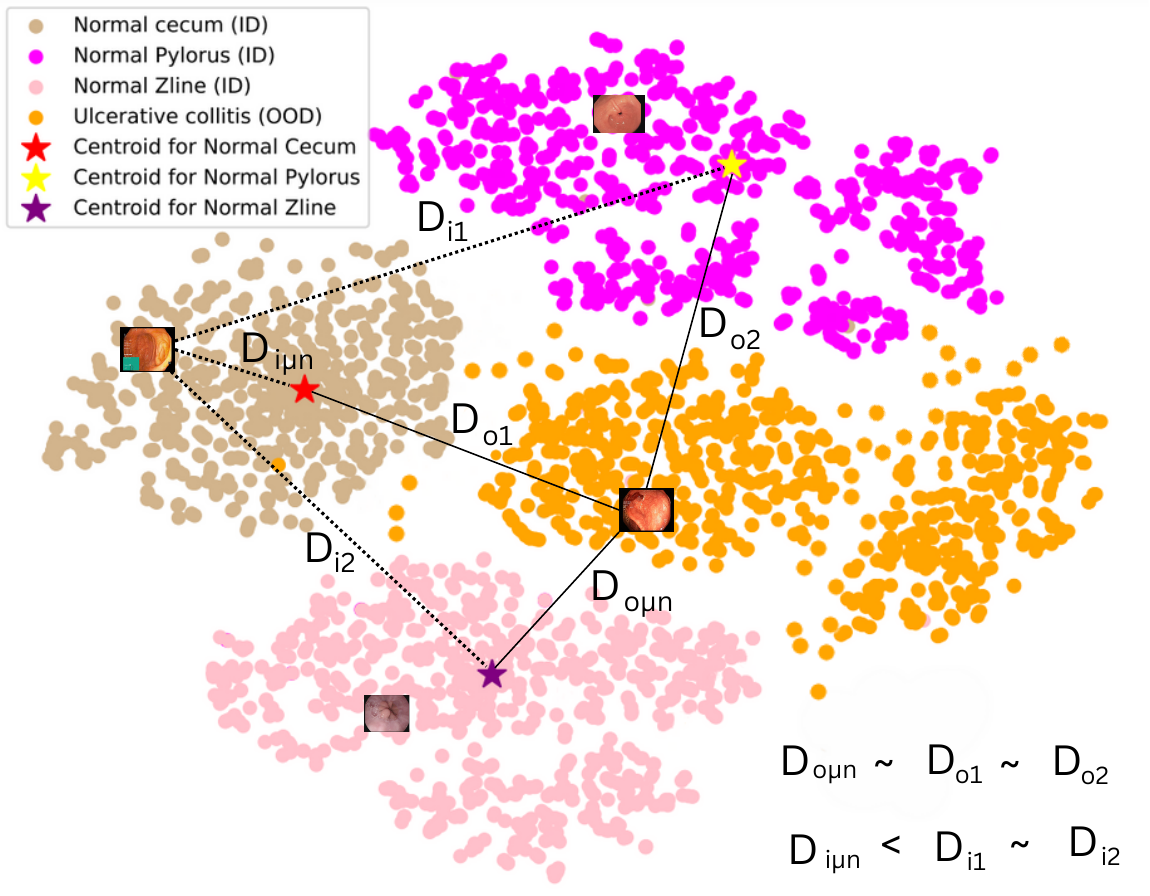}
\centering
\caption{t-SNE plot of Feature Space representation of the ViT model for the Kvasir dataset: In-Distribution data, Z-line (pink), Cecum (brown), and Pylorus (magenta) aligning with their respective centroid while OOD data, i.e. Ulcerative Colitis(orange) is scattered and at a distance from all In-Distribution centroids. The model pushes the ID centroids as far away as possible during training while the OOD data unseen at train time are more scattered in feature space. In essence, the nearest centroid distance for ID sample is significantly smaller compared to its distance from non-nearest centroids whereas for OOD data it is more or less similar.} 
\label{fig:tsne_plot}
\end{figure}

Yet another recent study on natural images~\cite{knn-ood} has shown that in-distribution (ID) data cluster around their respective class centroids, while out-of-distribution (OOD) data, when mapped onto the same feature space, generally remain distant from any specific ID centroid, reflecting its characteristics different from ID classes. As a result, an ID test sample exhibits a smaller nearest-neighbour distance to its closest cluster compared to an OOD test example.  However, this behaviour does not hold in the medical domain, where abnormalities labeled as OOD can exhibit characteristics similar to ID samples, making it a near-OOD detection case as we can see in a t-SNE plot of ID/OOD embeddings shown in Figure~\ref{fig:tsne_plot}. Thus, we argue that considering only the distance to the nearest centroid is not sufficient, and hence, we propose a scoring method that considers not only the distance of a test sample to the nearest cluster centroid but also the sum of its distances to other centroids, capturing a more nuanced representation of near-OOD instances.

OOD samples tend to have similar distances to their nearest ($D_{o\mu n}$) and non-nearest centroids ($D_{o1}, D_{o2}$) i.e. $D_{o\mu n} \sim D_{o1} \sim D_{o2}$. This is because OOD samples do not align strongly with any specific class clusters as they are unseen during training. In contrast, ID samples are naturally closer to their nearest cluster centroid ($D_{i \mu n}$) than to non-nearest centroids ($D_{i1}, D_{i2}$) i.e. $D_{i\mu n} < D_{i1} \sim D_{i2}$. This behaviour reflects the clustering effect created by training. 

Additionally, due to the intra-class attraction induced by cross-entropy ID samples exhibit smaller distances to their nearest centroid compared to OOD samples i.e. ($D_{i \mu n} < D_{o \mu n}$). At the same time, ID samples show larger distances to non-nearest centroids again attributed to the inter-class repulsion effect of the same loss function. OOD samples, being more dispersed around feature space, have a relatively smaller combined distance to non-nearest centroids ($D_{o1}+D_{o2}$). Figure~\ref{fig:tsne_plot} illustrates these patterns showing that the distances to non-nearest centroids for OOD samples ($D_{o1}+D_{o2}$) smaller compared to ID samples ($D_{i1} + D_{i2}$). These findings are consistent with the ablation study on distances seen in ID and OOD data presented in Table \ref{tab: distance_ablation}.

\subsection{Nearest Centroid Distance Deficit}

Based on our observation mentioned earlier, we propose a novel scoring method, Nearest Centroid Distance Difference (NCDD), to enhance the robustness of OOD detection. Our pipeline is illustrated in Figure \ref{pipelinefig} (b) taking the reference of Kvasir\cite{kvasir} dataset with three different normal anatomical landmarks as ID categories, and any type of anomalies as OOD. Following feature extraction, our approach comprises three key steps : (1) class-specific Centroid estimation of ID examples, (2) NCDD score calculation, and (3) OOD detection decision.  We discuss these steps in detail in the following sections. 

\subsubsection{Centroid estimation}
To compute a class-specific centroid, we average the feature representations of  examples belonging to a class, as defined by the following equation:  
\begin{equation} \label{eq2}
\begin{split}
\text{Class-wise centroid }: \quad \mu_c = \frac{1}{N_c} \sum_{i=1}^{N_c} z_i^c,
\end{split}
\end{equation}
where $N_c$ is the total number of samples belonging to class $ c $, and  $z_i^c$ represents the feature representation of the $ i $-th sample in class $c$ within the training dataset.

\subsubsection{NCDD}

%

Given a test image $x$, we obtain its feature representation as $z$. Then, the Euclidean distance to class centroids is calculated as $D_c =   \lVert \mu_c - z \rVert_2 $. We then define the distance of a test sample to its nearest cluster centroid as the nearest centroid distance $D_{\mu_n}$ i.e
\begin{equation}
D_{\mu_n} = argmin_c D_c
\end{equation}
and the sum of its distances to all other centroids as the non-nearest distance $D_{\mu_m}$ i.e.
\begin{equation}\label{non_nearest_eqn}
D_{\mu_m} = \sum_{c \neq \text{argmin}_c D_c} D_c
\end{equation}

Finally, our proposed OOD score for a test image is computed as the sum of its distances to non-nearest centroids, minus the distance to the nearest neighbour, 
\begin{equation} \label{eq3}
\begin{split}
\text{NCDD} =   \alpha \cdot D_{\mu_m} -  \beta \cdot 
D_{\mu_n},\\ 
\end{split}
\end{equation}

where $\alpha$ and $\beta$ represent the weight for each distance measure and $D_{\mu_m}$ as defined in \ref{non_nearest_eqn}. Motivated by the work of 
Zihan Zhang \textit{et al.} \cite{zhang2023decoupling}, which demonstrated that log of L1-norm of the penultimate layer feature, \textit{i.e.,} $\|\mathbf{z}\|_1$ = $\sum_{i=1}^{n} |z_i|$,  captures the notion of OOD detection, we incorporate this into the weight terms $\alpha$ and $\beta$ further to enhance the effectiveness of our proposed OOD score. Thus, $\alpha$ and $\beta$ are defined as:
\begin{equation} \label{eq31}
\alpha = \log\left(\frac{\|\mathbf{z}\|_1}{10^{\alpha_1}}\right) \text{and} \ \beta = \log\left(\frac{\|\mathbf{z}\|_1}{10^{\alpha_2}}\right)
\end{equation}
where $\alpha_1$ and $\alpha_2$ serve as hyperparameters, tuned during test time and based on the unique attributes of the datasets. By incorporating the logarithm, the weights $\alpha$ and $\beta$ become more sensitive to changes in the magnitude of the feature vector of individual OOD data, further scaling the importance of $D_{\mu_m}$ and $D_{\mu_n}$ accordingly. 
 
The decision for whether a test image is OOD or not can determined based on NCDD score. If the score is smaller than the threshold, estimated by cross-validation, it is OOD or ID example.  Algorithm~\ref{algo:NCDD_algo} shows the pseudocode to estimate NCDD. 

\begin{algorithm}
\caption{Algorithm to compute NCDD Score for OOD detection}
\begin{algorithmic}[1]
\REQUIRE Input image $I$, Model $M$ trained on landmark classification dataset, Threshold $\lambda$
\ENSURE Predicted class label $C$ (ID or OOD)

\STATE Extract feature vector $z$ from penultimate layer of model $M$
\STATE Calculate class-wise centroids for each class in the training set: $\mu_c = \frac{1}{N_c} \sum_{i=1}^{N_c} (z_i^c))$
\FOR{each test image $x$}
    \STATE $z$ = $\varphi(x)$ 
    \STATE Compute distances: $d_c = \|\mu_c - z\|_2$ for all classes $c$
    \STATE $D_{\mu n} \gets \min_c d_c$ \COMMENT{Distance to nearest centroid}
    \STATE $D_{\mu m} \gets \sum_{c \neq \text{argmin}_c d_c} d_c$ \COMMENT{Sum of distances to non-nearest centroids}
    
    \STATE $\text{NCDD} = \alpha \cdot D_{\mu m} - \beta \cdot D_{\mu n}$
    \IF{$\text{NCDD} \leq \lambda$}
        \STATE Classify as OOD
    \ELSE
        \STATE Classify as ID
    \ENDIF
\ENDFOR
\end{algorithmic}
\label{algo:NCDD_algo}
\end{algorithm}

\section{Experiments}
\subsection{Datasets and Implementation Details}

To test our method in real-world settings, we employed two multi-class endoscopy datasets designed for Gastrointestinal Disease Detection: Kvasir \cite{kvasir} and Gastrovision \cite{gastrovision}. These datasets include images from normal and anatomical findings and the most common abnormalities occurring in various upper and lower GI tract regions. We preprocessed this dataset into an OOD setting \cite{selfsupendo} and designated the classes related to normal findings and anatomical landmarks as in-distribution (ID) classes while segregating the remaining categories as out-of-distribution (OOD) classes.\\

\subsubsection{Datasets}
The \textbf{Kvasirv2}\cite{kvasir} dataset comprises 8 categories depicting anatomical landmarks, pathological findings, or endoscopic procedures within the gastrointestinal (GI) tract with 1000 images for each class. The anatomical landmarks in the Kvasirv2 dataset encompass the Z-line, Pylorus, and Cecum. At the same time, pathological findings include Esophagitis(ESO), Polyps(POL), and Ulcerative colitis(UC) and images related to polyp removal, namely the ``dyed and lifted polyp"(DLP) class and the ``dyed resection margins"(DRM). We designated the classes derived from anatomical landmarks in the Kvasirv2 dataset as in-distribution (ID) classes, treating the remaining classes as out-of-distribution representing unhealthy cases. We randomly selected 2400 in-distribution images for model training and employed 600 in-distribution images, combined with 5000 out-of-distribution images, to assess the final model's performance in OOD detection. 

The \textbf{Gastrovision}\cite{gastrovision} dataset comprises a total of twenty-seven categories, featuring 8,010 images obtained from examinations of both the upper and lower gastrointestinal (GI) tracts.  
The dataset is categorized into normal findings, anatomical landmarks(11 classes) and pathological findings, therapeutic interventions(16 classes). To facilitate model training, we randomly selected 3804 in-distribution images(normal findings and anatomical landmarks), utilizing a combination of 955 ID images and 3241 OOD images to comprehensively evaluate the final model's performance in OOD detection.\\

\subsubsection{Implementation Details}
We trained several wide ranges of models, including ResNet-18~\cite{resnet}, ViT-Small~\cite{vit}, DeiT-base~\cite{deit}, and MLP-Mixer-small~\cite{mlpmixer}, on both the Kvasirv2 and GastroVision datasets. For the Kvasirv2 dataset, Resnet-18, ViT-Small, DeiT-base and MLP-mixer-small models were trained for 20, 20, 50 and 50 epochs, respectively. And for the Gastrovision dataset, Resnet-18, ViT-Small, DeiT-base and MLP-mixer-small models were trained for 20, 50, 50 and 50 epochs, respectively. 

In all experiments, we used a batch size of 32 and the Adam optimizer, starting with an initial learning rate of 1×10$^{-4}$. All models were initialized with ImageNet pre-trained weights and employed the standard cross-entropy loss function. The input images were resized to 224x224 pixels. 

The models were trained and tested using PyTorch v2.1.0 on an NVIDIA A100 GPU. To tune the hyperparameters $\alpha_1$ and $\alpha_2$, we created synthetic validation OOD data. This OOD data is curated by randomly selecting a rectangle region in the image replacing the region with random values in ID validation data and then further adding speckle noise following Hendrycks \textit{et al.} \cite{hendrycks2018deep}.

\begin{table}[!t]
    \centering
    \caption{Quantitative metrics of the downstream task model used for OOD detection}
    \begin{tabular}{c c  c c c c}
        \hline
        \textbf{Model} & \multicolumn{1}{c}{\textbf{Parameters}} & \multicolumn{1}{c}{\textbf{Feat.Dim}} & \multicolumn{2}{c}
        {\textbf{Accuracy}}  \\
        &&&\textbf{Kvasirv2} & \textbf{Gastrovision} \\ \hline
        Resnet-18 & 11M & 512 & 98.57 & 83.66  \\
        ViT & 21M & 384 & 99.10 & 85.96 \\
        DeiT & 21M & 512 & 99.16 & 83.24  \\
        MLPmixer & 59M & 768 & 99.5 & 85.44 \\
        \hline
    \end{tabular} 
    \label{tab: Downstream_task}
\end{table}

\section{Results}
We assessed the efficacy of our method's OOD detection in gastrointestinal settings against established state-of-the-art OOD detection methods from existing literature, logit-based methods MSP \cite{msp}, ODIN \cite{Odin}, Energy \cite{energyscore}, Entropy \cite{entropy}, MaxLogit \cite{maxlogit}, and feature-based methods KNN \cite{knn-ood}. The model's performance for OOD evaluation was optimal for the respective downstream task, Table.\ref{tab: Downstream_task}.

\subsection{Quantative Results}
\begin{table}
    \centering
    \caption{Quantitative Comparison with other methods in terms of AUC$\uparrow$ and FPR95$\downarrow$ for Resnet-18 on Kvasirv2 and Gastrovision datasets. Bold characters denote the best performance and underlined characters denote the second-best performance for a metric.}
    \begin{tabular}{c c cc cc}
    \hline
    &
        &\multicolumn{2}{c}{\textbf{Kvasirv2}} & \multicolumn{2}{c}{\textbf{Gastrovision}} \\
        
        \textbf{Method}& \textbf{Venue}& \textbf{AUC} $\uparrow$ & \textbf{FPR95}$\downarrow$ & \textbf{AUC}$\uparrow$& \textbf{FPR95}$\downarrow$ \\
        \hline
        
        MSP & ICLR'17 & \textbf{87.57} & 33.06 & 69.82 & 87.2 \\
        ODIN & ICLR'18 & \underline{86.95} & 40.48 & \underline{74.12} & \underline{79.54} \\ 
        Energy & NeurIPS'20 & 86.00 & 43.28 & 73.11 & 85.25 \\
        Entropy & IJCNN'21& 87.55 & \underline{33.02} & 70.53 & 85.81 \\
        MaxLogit & ICML'22 & 86.04 & 43.04 & 72.95 & 84.91 \\
        KNN & ICML'22 & 86.59 & 36.02 & 71.37 & 83.62 \\
        BLOOD & ICLR'24 & 52.1 & 93.56 & 53.11 & 92.88 \\
        Neco & ICLR'24 & 86.29 & 41.88 & 73.31 & 84.57 \\
        FDBD & ICML'24 & 84.77 & 36.88 & 72.63 & 80.93 \\
        NCDD (Ours) & - & 85.68 & \textbf{30.86} & \textbf{79.1} & \textbf{72.6} \\
        \hline
    \end{tabular}
    \label{tab:resnet18_results}
\end{table}

\subsubsection{Metrics} 
\textbf{AUC} and \textbf{FPR95} measures were used to quantify the performance of our OOD detection method. AUC measures the area under the receiver operating characteristic curve, with higher values indicating better performance. FPR95 represents the false positive rate when the true positive rate is 95\%, with smaller values indicating better performance. Low FPR95 is essential in the OOD detection task as it reduces the false positive rate, ultimately enhancing the trustworthiness of the model.

Results in Table \ref{tab:resnet18_results}, compares our method with existing SOTA methods on Resnet18 with feature dimension 512. The generalizability and model-agnostic nature of our method is supported by the fact that NCDD consistently outperforms established OOD methods in both AUC and FPR95 scores across various model architectures like ViT (table \ref{tab:vit_results}), MLP-Mixer (table \ref{tab:MLPMixer}) and DEiT (table \ref{tab:Deit_results}).

Also, from Figure \ref{bargraph}, we can see that our method increases the distinction capacity of the model on individual OOD classes, i.e. on separate cases of abnormalities. The Esophagitis OOD class shares significant feature overlap with the Normal Z-line class in training (ID) data, leading to a decline in the performance of feature-based OOD detection methods when it comes to identifying Esophagitis.  In cases where the distinction is noticeable, and the model can capture appropriate feature information, we can see significant improvement in OOD detection by NCDD over other methods. Its consistent performance across different architectures and diverse datasets makes it promising for OOD detection in endoscopic images, enhancing the trustworthiness of an AI-assisted procedure in diagnosis. 

\begin{table}
    \centering
    \caption{Quantitative Comparison with other methods regarding AUC$\uparrow$ and FPR95$\downarrow$ for ViT-Small on Kvasirv2 and Gastrovision datasets. Bold characters denote the best performance, and underlined characters denote the second-best performance for a metric.}
    \begin{tabular}{c c cc cc}
    \hline
    &
        &\multicolumn{2}{c}{\textbf{Kvasirv2}} & \multicolumn{2}{c}{\textbf{Gastrovision}} \\
        
        \textbf{Method}& \textbf{Venue}& \textbf{AUC}$\uparrow$ & \textbf{FPR95}$\downarrow$ & \textbf{AUC}$\uparrow$ & \textbf{FPR95}$\downarrow$ \\
        \hline
        
        MSP & ICLR'17 & 85.93 & 39.74 & 75.08 & 84.36 \\
        ODIN & ICLR'18 & 83.84 & 42.10 & \underline{79.36} & \underline{68.40} \\ 
        Energy & NeurIPS'20 & 83.52 & 43.12 & 78.62 & 73.62 \\
        Entropy & IJCNN'21& 85.87 & 39.6 & 75.50 & 81.06 \\
        MaxLogit & ICML'22 & 83.53 & 43.12 & 78.45 & 75.53 \\
        KNN & ICML'22 & \underline{90.74} & \underline{38.44} & 78.43 & 80.87 \\
        BLOOD & ICLR'24 & 51.61 & 93.10 & 52.02 & 93.37 \\
        Neco & ICLR'24 & 87.87 & 27.8 & 80.16 & 75.53 \\
        FDBD & ICML'24 & 88.31 & 29.0 & 75.75 & 75.81 \\
        NCDD (Ours) & - & \textbf{91.61} & \textbf{22.94} & \textbf{83.95} & \textbf{55.63} \\
        \hline
    \end{tabular}
    \label{tab:vit_results}
\end{table}

\begin{table}
    \centering
    \caption{Quantitative Comparison with other methods regarding AUC$\uparrow$ and FPR95$\downarrow$ for DeiT on Kvasirv2 and Gastrovision datasets. Bold characters denote the best performance and underlined characters denote the second-best performance for a metric.}
    \begin{tabular}{c c cc cc}
    \hline
    &
        &\multicolumn{2}{c}{\textbf{Kvasirv2}} & \multicolumn{2}{c}{\textbf{Gastrovision}} \\
        
        \textbf{Method}& \textbf{Venue}& \textbf{AUC}$\uparrow$ & \textbf{FPR95}$\downarrow$ & \textbf{AUC}$\uparrow$ & \textbf{FPR95}$\downarrow$ \\
        \hline
        
        MSP & ICLR'17 & 74.57 & 52.26 & 69.53 & 86.05 \\
        ODIN & ICLR'18 & 74.15 & \underline{52.18} & 77.48 & 72.69 \\ 
        Energy & NeurIPS'20 & 72.92 & 53.82 & 75.91 & 75.56 \\
        Entropy & IJCNN'21 & 74.56 & 52.38 & 72.49 & 79.39 \\
        MaxLogit & ICML'22 & 72.92 & 53.82 & 75.0 & 79.17 \\
        KNN & ICML'22 & \underline{82.9} & 52.9 & \textbf{85.69} & \underline{54.71} \\
        BLOOD & ICLR'24 & 52.65 & 93.92 & 53.16 & 93.09 \\
        Neco & ICLR'24 & 68.45 & 66.88 & 78.67 & 72.45 \\
        FDBD & ICML'24 & 71.01 & 53.84 & 76.01 &  74.45\\
        NCDD (Ours) & - & \textbf{87.09} & \textbf{37.9} & \underline{85.37} & \textbf{52.48} \\
        \hline
    \end{tabular}
    \label{tab:Deit_results}
\end{table}

\begin{table}
    \centering
    \caption{Quantitative Comparison with other methods regarding AUC$\uparrow$ and FPR95$\downarrow$ for MLP-Mixer on Kvasirv2 and Gastrovision datasets. Bold characters denote the best performance, and underlined characters denote the second-best performance for a metric.}
    \begin{tabular}{c c cc cc}
    \hline
    &
        &\multicolumn{2}{c}{\textbf{Kvasirv2}} & \multicolumn{2}{c}{\textbf{Gastrovision}} \\
        
        \textbf{Method}& \textbf{Venue}& \textbf{AUC}$\uparrow$ & \textbf{FPR95}$\downarrow$ & \textbf{AUC}$\uparrow$ & \textbf{FPR95}$\downarrow$ \\
        \hline
        
        MSP & ICLR'17 & 78.66 & 48.54 & 72.35 & 88.12 \\
        ODIN & ICLR'18 & 78.19 & 47.28 & 78.17 & \underline{70.26} \\ 
        Energy & NeurIPS'20 & 70.70 & 54.86 & 77.31 & 72.66 \\
        Entropy & IJCNN'21 & 78.66 & 48.14 & 74.31 & 81.06 \\
        MaxLogit & ICML'22 & 70.70 & 54.86 & 76.44 & 77.29 \\
        KNN & ICML'22 & \underline{88.77} & \underline{31.80} & \underline{78.71} & 74.27 \\
        BLOOD & ICLR'24 & 50.66 & 94.58 & 53.36 & 93.40 \\
        Neco & ICLR'24 & 73.81 & 52.1 & 78.45 & 73.31 \\
        FDBD & ICML'24 & 75.39 & 49.96 & 73.89 & 77.14 \\
        NCDD (Ours) & - & \textbf{92.44} & \textbf{21.92} & \textbf{80.93} & \textbf{60.5} \\
        \hline
    \end{tabular}
    \label{tab:MLPMixer}
\end{table}

\subsection{Qualitative Results}

\begin{figure*}[t]
\includegraphics[width=\linewidth]{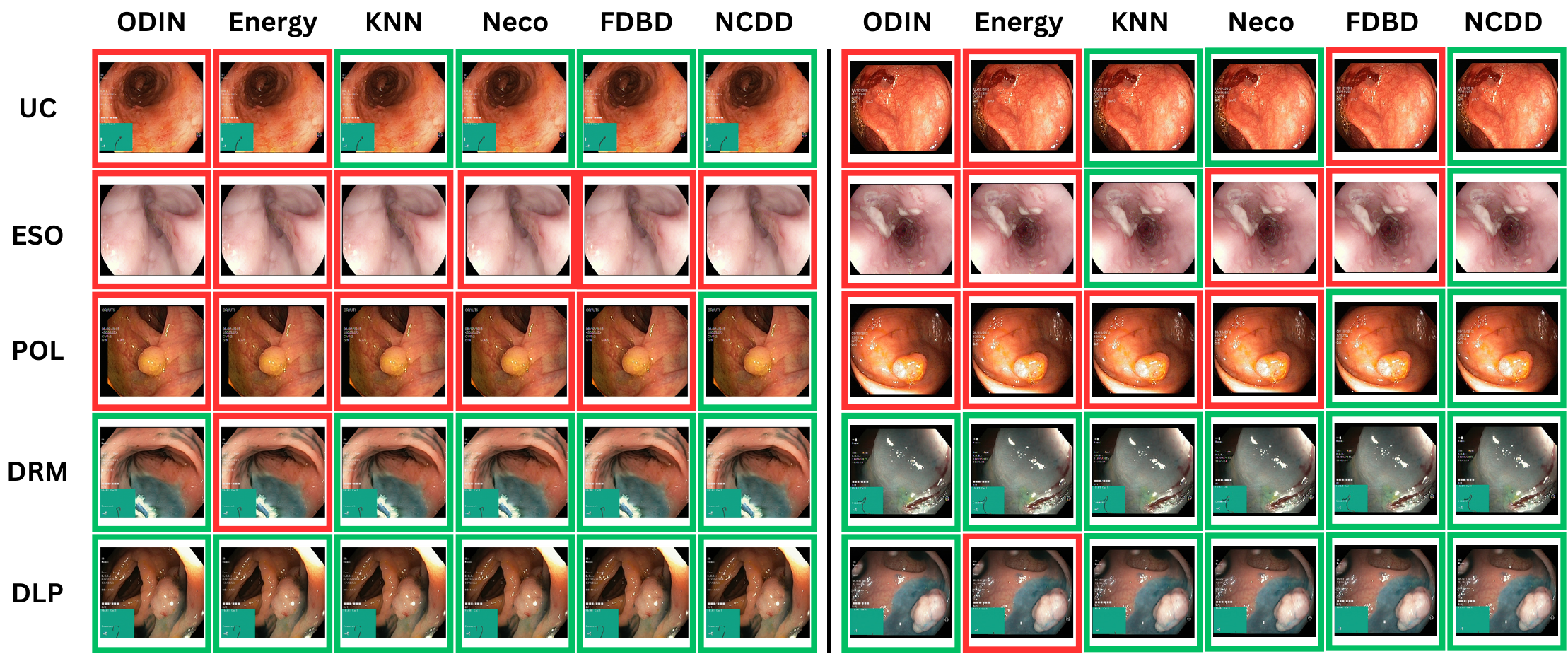}
\caption{Qualitative comparison of our method and other SOTA OOD methods for Kvasirv2 on ViT model: OOD examples over-confidently predicted by the corresponding method as healthy ID data are indicated inside red frame, while images correctly identified as OOD(abnormality) are indicated in green.} \label{qualifig}
\end{figure*}

\begin{figure}
\includegraphics[width=1\linewidth]{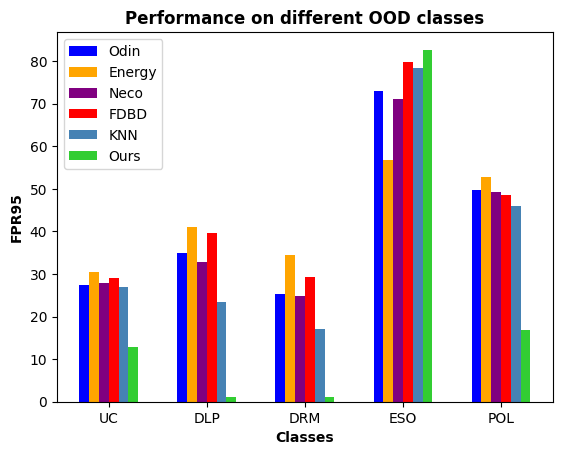}
\caption{Performance of our method on FPR95$\downarrow$ on separate te OOD classes for Kvasirv2 on ViT model. It is evident that, out of four OOD examples, NCDD outperforms KNN-OOD by a significant margin.} \label{bargraph}
\end{figure}

The qualitative performance comparison of our proposed method, NCDD, with popular OOD methods like KNN-OOD, Neco, and FDBD, as presented in Figure \ref{qualifig}, clearly demonstrates the superior robustness of NCDD across various OOD categories. Our method consistently outperforms other methods, particularly in challenging scenarios where existing methods struggle.

For instance, in the case of severe OOD examples like Polyps (POL), which pose significant classification challenges, NCDD exhibits remarkable accuracy. While methods like KNN-OOD and Neco often misclassify these complex cases as in-distribution, our method successfully identifies these samples as OOD. 

\subsection{Ablation Studies}
Ablation experiments were conducted to investigate the impact of various distance terms on our out-of-distribution (OOD) score performance. The results presented in Table \ref{tab: distance_ablation} and \ref{tab: hyperparams_ablation} demonstrate that while utilizing only the nearest centroid distance yields better performance than considering both nearest centroid distance and non-nearest centroid distance, achieving optimal performance requires appropriately scaling both terms.

\begin{table}[htbp]
    \centering
    \caption{Significance of $D_{\mu_m}$ and $D_{\mu_n}$ for OOD scoring for Resnet-18 on Kvasirv2 dataset.}
    \begin{tabular}{c c c c c}
        \hline
        \textbf{Distance} & \multicolumn{1}{c}{\textbf{Mean}} & \multicolumn{1}{c}{\textbf{Mean}} & \multicolumn{1}{c} {\textbf{FPR}$\downarrow$} \\ 
        \textbf{score} & \multicolumn{1}{c}{\textbf{ID-Score}} & \multicolumn{1}{c}{\textbf{OOD-Score}} & \\ \hline
        $D_{\mu_m}$ & 5.82 & 5.36 & 33.60  \\
        $-D_{\mu_n}$ & $-$0.16 & $-$0.64  & 33.50  \\
        $D_{\mu_m} - D_{\mu_n}$ & 5.66 & 4.72 & 32.42  \\     $\alpha \cdot D_{\mu_m} -  \beta \cdot D_{\mu_n}$ & 47.95 & 40.01 & \textbf{30.86} \\   
        \hline
    \end{tabular} 
    \label{tab: distance_ablation}
\end{table}

\begin{table}[t!]
    \centering
    \caption{Hyperparameter study for NCDD on Resnet-18 with synthetic validation OOD obtained from Kvasirv2 dataset following Hendrycks \textit{et al.}\cite{hendrycks2018deep}}
    \begin{tabular}{c c}
        \hline
        \textbf{Condition} & \multicolumn{1}{c} {\textbf{FPR}$\downarrow$} \\ \hline
        $\alpha_1 = -1$ \& $\alpha_2 = -1$  & 29.67 \\
        $\alpha_1 = -1$ \& $\alpha_2 = 0$  & \textbf{28.83}  \\
        $\alpha_1 = -1$ \& $\alpha_2 = 1$ & 29.00  \\
        $\alpha_1 =0 $ \& $\alpha_2 = 0$  & 29.17  \\
         $\alpha_1 = -2$ \& $\alpha_2 = 0$  & 29.67  \\
        \hline
    \end{tabular} 
    \label{tab: hyperparams_ablation}
\end{table}

\section{Conclusion}
We reformulate abnormality detection in gastrointestinal images as an OOD problem to enable the recognition of abnormalities in the GI tract through a model trained only on abundant normal anatomical findings data. We introduced a novel distance-based OOD detection method, Nearest Centroid Distant Deficit (NCDD), which utilizes the feature space distance of ID data in multi-class gastrointestinal image datasets and its relative positioning in feature space to discern them from OOD samples. NCDD outperforms the previous state-of-the-art OOD detection methods in gastrointestinal vision tested in two datasets on AUC and FPR95 metrics over four model architectures. 

Overall, we demonstrate tremendous potential for supervised models on OOD detection as a general-purpose tool for flagging abnormalities in endoscopy images from gastrointestinal settings, an unexplored area in medical imaging research. The clinical applicability of this method is most promising in the current medical context as it is relatively easy to integrate and also allows for expert intervention, unlike fully autonomous AI methods.

\bibliographystyle{ieeetr}
\bibliography{ref}

\end{document}